%% file: main.tex
\documentclass[letterpaper, 10 pt, conference]{ieeeconf}   %
\IEEEoverridecommandlockouts                              %

\input{math_commands.tex}

\usepackage{graphicx}
\usepackage{cite}
\usepackage[colorlinks,linkcolor=red]{hyperref}
\usepackage{url}
\usepackage{booktabs}
\usepackage{array}
\usepackage{adjustbox}
\usepackage{makecell}
\usepackage{algorithm}
\usepackage{algorithmic}
\usepackage{cleveref}

\usepackage{dblfloatfix}
\usepackage{xspace}

\makeatletter
\DeclareRobustCommand\onedot{\futurelet\@let@token\@onedot}
\def\@onedot{\ifx\@let@token.\else.\null\fi\xspace}

\def\etal{\emph{et al}\onedot}
\makeatother

\title{\LARGE \bf
Learning Human-Aware Robot Policies for Adaptive Assistance
}

\author{Jason Qin$^{1}$, Shikun Ban$^{2}$, Wentao Zhu$^{2\dagger}$, Yizhou Wang$^{2}$, and Dimitris Samaras$^{1}$%
\thanks{$^{1}$First Author and Fifth Author are with Department of Computer Science, Stony Brook University, USA
{\tt\small jaqin@cs.stonybrook.edu, samaras@cs.stonybrook.edu},
}%
\thanks{$^{2}$Second Author, Third Author, and Forth Author are with Center on Frontiers of Computing Studies, Peking University, China {\tt\small bansk@stu.pku.edu.cn, wtzhu@pku.edu.cn, yizhou.wang@pku.edu.cn},
}%
\thanks{$^{\dagger}$ Corresponding author.}
}

\begin{document}

\maketitle
\thispagestyle{empty}
\pagestyle{empty}

\begin{abstract}
Developing robots that can assist humans efficiently, safely, and adaptively is crucial for real-world applications such as healthcare. While previous work often assumes a centralized system for co-optimizing human-robot interactions, we argue that real-world scenarios are much more complicated, as humans have individual preferences regarding how tasks are performed. 
Robots typically lack direct access to these implicit preferences. However, to provide effective assistance, robots must still be able to recognize and adapt to the individual needs and preferences of different users.
To address these challenges, we propose a novel framework in which robots infer human intentions and reason about human utilities through interaction. 
Our approach features two critical modules:
the \emph{anticipation module} is a motion predictor that captures the spatial-temporal relationship between the robot agent and user agent,
which contributes to predicting human behavior; 
the \emph{utility module} infers the underlying human utility functions through progressive task demonstration sampling. 
Extensive experiments across various robot types and assistive tasks demonstrate that the proposed framework not only enhances task success and efficiency but also significantly improves user satisfaction, paving the way for more personalized and adaptive assistive robotic systems.
Code and demos are available at \url{https://asonin.github.io/Human-Aware-Assistance/}.

\end{abstract}

\input{sections/1Intro}

\input{sections/2Related}

\input{sections/3Method}

\input{sections/4Experiments}

\input{sections/5Conclusion}

\bibliographystyle{IEEEtran}
\bibliography{IEEEabrv,iclr2025_conference}

\input{sections/6Appendix}

\end{document}

%% file: math_commands.tex
\usepackage{amsmath,amsfonts,bm}

\def\eqref#1{equation~\ref{#1}}

\def\1{\bm{1}}

\DeclareMathAlphabet{\mathsfit}{\encodingdefault}{\sfdefault}{m}{sl}
\SetMathAlphabet{\mathsfit}{bold}{\encodingdefault}{\sfdefault}{bx}{n}

%% file: sections/1Intro.tex
\section{Introduction}

\begin{figure}[thpb]
  \centering
  \includegraphics[width=0.9\columnwidth]{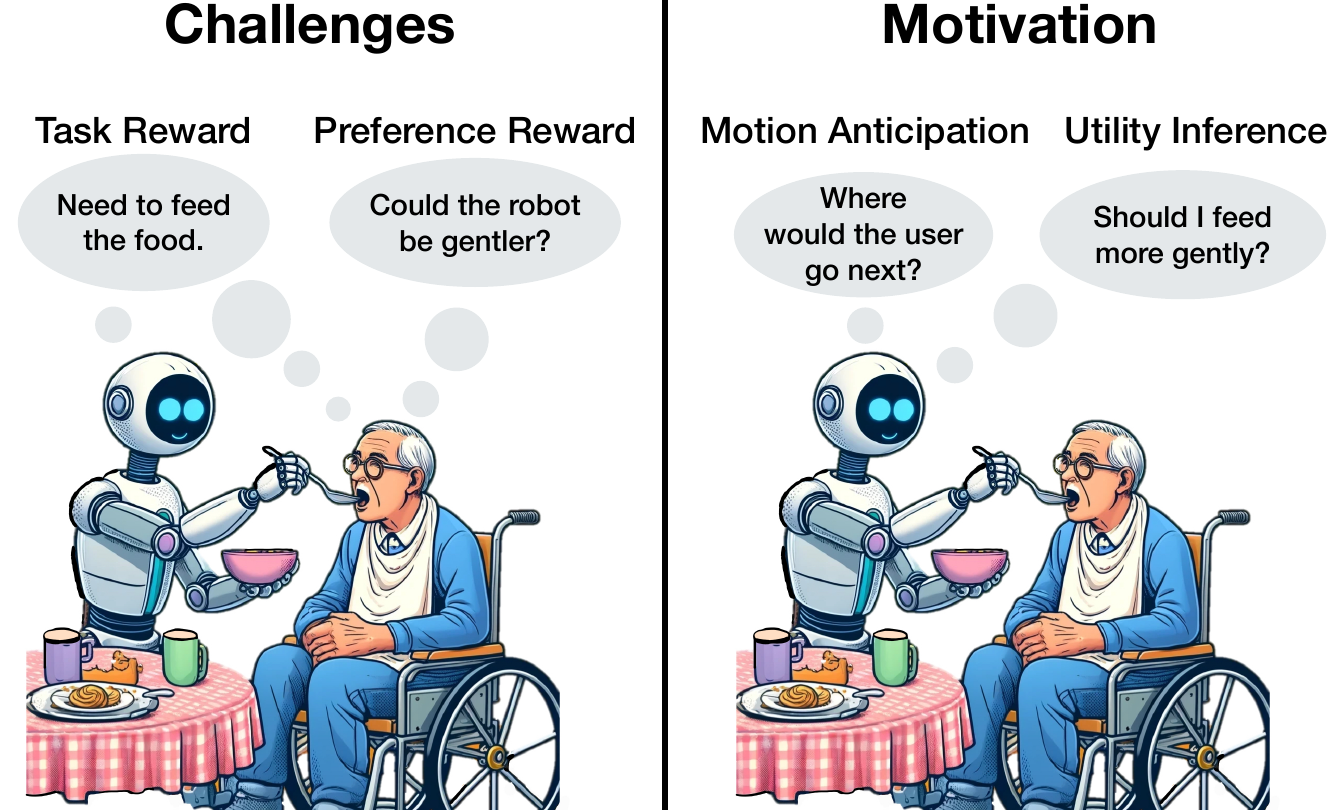}
  \caption{
  In our task scenario (demonstrated with the feeding example in the figure), the robot's initial objective is to achieve the basic goal of "feeding the food," represented by the \emph{Task Reward}. However, as shown in \textbf{Left}, the human user also has more nuanced \emph{Preference Reward}, which are unknown to the robot, leading to a misalignment between the human and robot reward functions.
To address this, we propose a novel framework, as depicted in \textbf{Right}. Beyond learning to fulfill the basic task requirements, we introduce two additional modules to better model human behavior and preferences: a \emph{Motion Anticipation} module for predicting the human's future motion and a \emph{Utility Inference} module for estimating user preferences.
}
  \label{fig:teaser}
  \vspace{-0.2cm}
\end{figure}

Developing robots that understand and assist humans is a critical long-term goal in Artificial Intelligence (AI) research. Beyond merely completing tasks, these systems must interact with humans in a friendly manner, dynamically coordinating and adapting to their needs. Such capabilities are essential for applications in home environments, assistive technologies, and healthcare. However, significant challenges remain, including ensuring safety, addressing the opacity of human goals, and navigating the diversity of human preferences.
For example, when assisting disabled or elderly individuals with feeding, users may prefer robots not only to deliver food to the mouth but also to handle it gently and avoid spills. 
These tasks, though seemingly simple, are inherently complex, 
requiring precision, safety, comfort, and adaptability to individual preferences.

To achieve this goal, previous research has explored human-robot collaboration in assistive scenarios
~\cite{erickson2020assistive,RCareWorld}
. 
Some studies
~\cite{erickson2020assistive,he2023learning,osa2024robustifying} 
propose modeling humans and robots as a unified system, using co-optimization to learn assistive policies where both agents are jointly trained with the same reward signal. 
However, this approach does not fully capture real-world dynamics. In practice, robots cannot fully access the human mind, leading to a gap or misalignment between human and robot reward functions. 
For the robot, the most straightforward \emph{task reward} is completing the task, such as whether the food is delivered to the intended location. For humans, there is an additional layer of \emph{preference reward} — preferences regarding how the task is performed, such as the speed or force used during food delivery~\cite{erickson2020assistive,liu2022task}. These human rewards are often difficult to articulate and vary across individuals, making them unknown to the robot.
Therefore, robots must gather information during interactions, inferring human intentions from behavioral cues and reasoning about the underlying human preferences~\cite{hoffman2024inferring,zhu2024language,jain2018recursive}. By adjusting their policies accordingly, robots can more effectively collaborate with humans to complete tasks while better aligning with human preferences and needs, thereby enhancing safety, robustness, and user satisfaction.

In light of this, we propose a more realistic formulation for assistive robotics and introduce a novel policy learning approach that first infers human intentions and utility functions, and then provides assistance accordingly. We first define an assistive task setting in which humans and robots cooperate under different reward signals. Specifically, robots are guided solely by clearly defined task rewards, while human behavior is driven by both task rewards and human preference rewards. This setting more accurately reflects real-world conditions and presents significant challenges.
To facilitate effective human-robot collaboration, we introduce a policy learning framework which comprises two key components, as shown in ~\Cref{fig:teaser}. 
Specifically, we propose an \emph{anticipation module} which enables the robot to predict human actions before determining its own actions. 
Furthermore, our \emph{utility module} estimates human utility function weight progressively from online interactions through distribution sampling, further improving the robot's adaptation ability to user preferences without the need of any preference queries.

We conduct comprehensive experiments across multiple tasks, preference settings, and robot types. Experimental results show that these modules not only improve task success and efficiency but also significantly enhance human rewards, resulting in a safer and more user-friendly experience. 
Our contributions are summarized as follows: (1) We propose a more realistic formulation in which human and robot rewards are not fully aligned. (2) We introduce an anticipation module, enabling the robot to predict human behavior before making its own action decisions. (3) We develop a utility module that infers the human utility function on the fly, without relying on explicit preference queries or human feedback.

%% file: sections/2Related.tex
\vspace{-0.2cm}
\section{Related Work}

\vspace{-0.1cm}
\subsection{Assistive Robotics}

Assistive robots have been explored as a viable solution to help disabled and elderly individuals perform daily tasks~\cite{chen2013robots, rhodes2018robot, nanavati2023design, bhattacharjee2019community, park2018multimodal, gallenberger2019transfer, erickson2020assistive, RCareWorld, he2023learning, osa2024robustifying,thumm2024human,zhou2023clothesnet, liu2022task}. Some works have focused on designing and developing virtual environments to facilitate the development and comparison of assistive algorithms~\cite{erickson2020assistive,RCareWorld,shen2021igibson}. 
For example, Assistive Gym\cite{erickson2020assistive} is a simulation framework for physical human-robot interaction and robotic assistance, providing several tasks crucial for daily living and introducing the co-optimization paradigm. Similarly, RCareWorld\cite{RCareWorld} proposes a framework for developing and evaluating socially-aware robotic assistance.
Researchers have also designed various approaches to enhance assistive performance~\cite{muller2016robot,unhelkar2018human,zhang2022learning,gallenberger2019transfer,jakhotiya2022improving,osa2022discovering,he2023learning,osa2024robustifying,madan2022sparcs,puthuveetil2022bodies, liu2022task}. For example, M{\"u}ller \etal\cite{muller2016robot} focus on integrating human-robot collaboration to enhance ergonomics, flexibility, and quality in the assembly line, 
Unhelkar \etal~\cite{unhelkar2018human} introduce a human-aware robotic assistant system that combines human motion prediction with time-optimal path planning.
Gallenberger \etal~\cite{gallenberger2019transfer} examine manipulation strategies for robotic feeding, focusing on how different bite acquisition and transfer methods impact the success and ease of feeding for users.
Jakhotiya \etal~\cite{jakhotiya2022improving} 
explores using recurrent neural networks to enhance policy learning for assistive tasks like feeding and dressing.
Osa \etal~\cite{osa2022discovering} introduce a solution by maximizing mutual information between state-action pairs and latent variables, allowing for a wider range of adaptive behaviors in tasks requiring robustness.
He \etal~\cite{he2023learning}
aims to efficiently transfer knowledge gained from one task to another by learning robust representations, enhancing the effectiveness and versatility of assistive robots.
Osa \etal~\cite{osa2024robustifying} 
suggests employing a latent variable to model the behavior of an opposing agent.

\subsection{Human-aware Policy Learning}

To develop collaborative robots better suited to work alongside or for humans, some works have proposed incorporating human feelings and experiences into consideration~\cite{christen2023learning,sadigh2017active,chen2022asha,palan2019learning,clegg2020learning,chao2022handoversim,liu2022task}.
Sadigh \etal~\cite{sadigh2017active} 
propose an interactive framework that efficiently learns complex reward landscapes from minimal user input, making the learning process both efficient and aligned with human preferences while the expressiveness of such minimal user input remains limited. 
Chen \etal~\cite{chen2022asha} present a human-in-the-loop approach for tele-operation in assistive robotics. While leveraging a control loop, their method enhances the robot's capability to deliver real-time assistance in dynamic environments.
Palan \etal~\cite{palan2019learning} 
enables systems to understand and predict what users value without them having to specify it explicitly each time.
Another line of works uses preference-based reinforcement learning (PbRL) to train agents with pairwise human feedback~\cite{NIPS2012_16c222aa, pbrl2013, wirth2017survey}. Recent advances aim to alleviate the reliance on explicit human feedback by introducing off-policy algorithms that relabel history experience and conduct unsupervised pre-training~\cite{Lee2021PEBBLEFI}, and training a preference predictor to provide pseudo preference labels~\cite{Zhan2020HumanGuidedRB, Cao2021WeakHP, Park2022SURFSR}.
Specifically, Liu  \etal~\cite{liu2022task} enhances the efficiency and robustness of preference-based reinforcement learning in complex interactive tasks by decoupling task objectives from preferences and integrating task prior knowledge into the PbRL framework. Their combination of the sketchy task reward shaping with an automated scripted teacher, reduces reliance on real human feedback during the training process.
Although these methods make valuable contributions to incorporating human factors into policy learning, they still often rely on human demonstrations or explicit preference queries, which limits their ability to autonomously learn and adapt to the complexity of human behavior. %

%% file: sections/3Method.tex
\begin{figure*}[t]
  \centering
  \includegraphics[width=1.8\columnwidth]{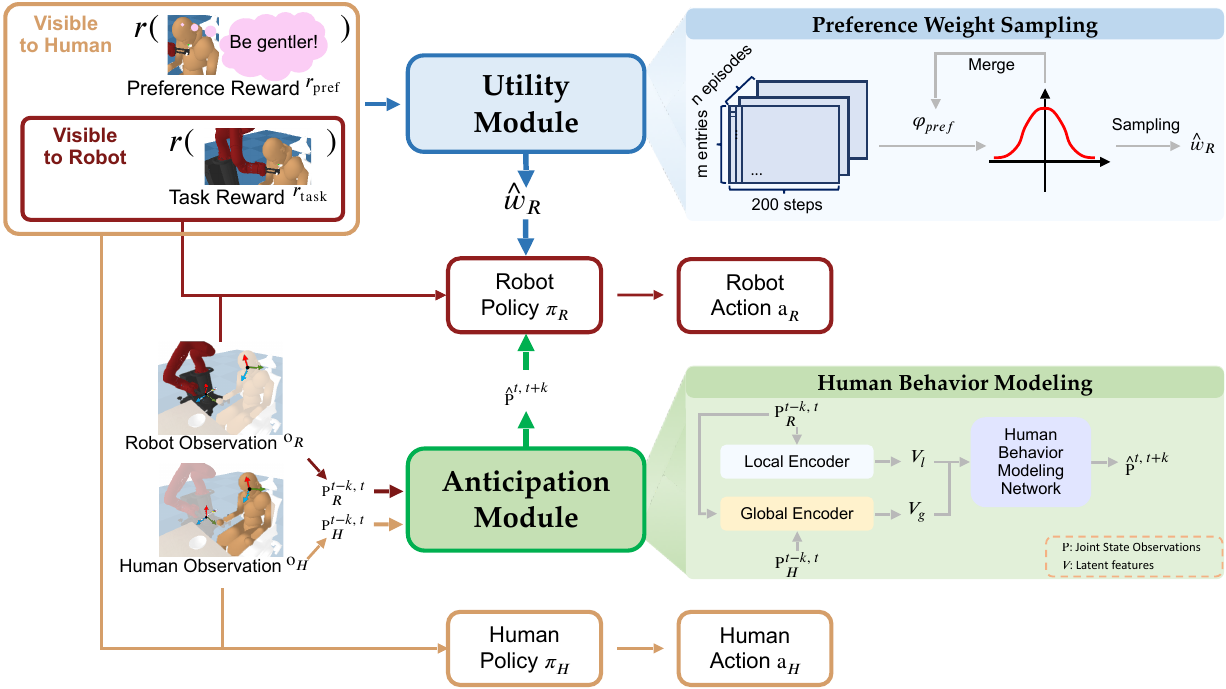}
  \caption{Overview of the proposed framework. 
  Each agent in the system is receiving an observation about its own information and some critical information about the other agent.
  The human agent is controlled by an independent policy powered by RL algorithms. 
  The robot agent consists of three parts: an RL backbone, an anticipation module~\ref{subsec:anticipation}, and a utility module~\ref{subsec:preference}.
  The \emph{anticipation module} predicts future human motion by taking past $k$ frames of joint information \(p_R^{t-k,t}\) and \(p_{H}^{t-k,t}\) from both agents, and predicting an anticipated  k steps future human joint information \(\hat{p}_{H}^{t,t+k}\).
The \emph{utility module} leverages the interaction histories to estimate a robot preference reward weight \( \hat{w}_H \), which is used to compute an estimated preference reward \( \hat{r}_{\text{pref}} \) to further guide robot policy learning.
  }
  \label{fig:framework}
    \vspace{-0.3cm}
\end{figure*}

\section{Method}
\label{sec:method}

In this section, we first introduce the problem setup and formulate the reward misalignment problem between human and robot. Then, we describe our proposed method step-by-step. The overview of the proposed human-aware policy learning framework is demonstrated in Figure \ref{fig:framework}.

\subsection{Problem Formulation}

The assistive task is modeled as a two-agent, finite-horizon decentralized partially observable Markov decision process (Dec-POMDP), defined by the tuple $(\mathcal{S}, \mathcal{O}, \mathcal{A}, T, r)$. Here, $\mathcal{S}$ is the state space, $\mathcal{A} = \mathcal{A}_H \times \mathcal{A}_R$ is the joint action space of the human ($\mathcal{A}_H$) and robot ($\mathcal{A}_R$), and $T(\mathbf{s}^{t+1} | \mathbf{s}^t, \mathbf{a}^t)$ defines the transition dynamics.
$\mathcal{O} = \mathcal{O}_H \times \mathcal{O}_R$ is the joint observation space, where the agents' observations are derived from the previous state and actions. The policies $\pi_H$ and $\pi_R$ generate actions based on the agents' observations, $\pi(a^t | o^t)$.
The reward function $r(\mathbf{s}, \mathbf{a})$ assigns a scalar reward based on the current state and actions. A trajectory \(\xi = \left( s^t, o_H^t, o_R^t, a_H^t, a_R^t \right)_{t=0}^N\) represents the sequence of states, observations, and actions over \(N\) timesteps.

Specifically, for a given trajectory $\xi$, we define two types of rewards: the task success reward and the human preference reward. The task reward is defined as:
\begin{equation}
r_\text{task} = \phi_\text{task}(\xi),
\label{eq:extrinsic_reward}
\end{equation}
where $\phi_\text{task}$ is a function that maps the agents' trajectories to a reward based on task completion and performance.
Meanwhile, the human preference reward is defined as:
\begin{equation}
r_\text{pref} = \phi_\text{pref}(\xi; w),
\label{eq:intrinsic_reward}
\end{equation}
where $\phi_\text{pref}(\xi; w)$ is a feature function parameterized by the weight parameter $\mathbf{w}$, mapping the trajectory $\xi$ to a scalar reward representing human satisfaction. The parameter $\mathbf{w}$ reflects the weight assigned to individual human preferences. In this work, we use a fixed set of human preference items and various preference settings. Each setting assigns different weights to the preference items, shaping the overall human preference reward.

We note that a fundamental misalignment often arises because robots cannot fully access or understand human preferences. 
These task performance preferences are subjective, vary among individuals, and are often opaque to robots.
Specifically, the human reward, combining both task and preference components, is:
\begin{equation}
    r_H(\xi) = r_\text{pref} + r_\text{task}
    \label{eq:human_reward}
\end{equation}
. In contrast, robots generally have access only to the task reward, with their objective focused solely on task success, \( r_R(\xi) = r_\text{task} \). This often leads to suboptimal outcomes from the human's perspective, highlighting the challenge of designing robots that can better incorporate human preferences for more satisfying interactions.

\subsection{Anticipation Module}
\label{subsec:anticipation}

To create robots that effectively integrate human preferences for more engaging interactions, 
an intuitive approach is to predict human future actions before making its own decisions~\cite{zhu2024social}. By using this predictive information to guide robot policies, the robot can generate movements that align more closely with human motion trends, reducing discomfort and enhancing task performance.
Given that agents' behavior patterns can be directly learned through observation~\cite{pomerleau1991efficient} within the limited observability, we attempt to establish an anticipation module $\mathcal{M}$ to predict human's joint positions $\hat{p}_H$ in future $k$ steps $\hat{p}_H^{t,t+k}$, as shown on the bottom of \Cref{fig:framework}. Specifically,
\begin{equation}
\hat{p}_{H}^{t,t+k} = \mathcal{M}(p_R^{t-k,t}, p_{H}^{t-k,t}), 
\label{eq:social_motion_model}
\end{equation}
where \(p_{H}^{t-k,t}\) and \(p_{R}^{t-k,t}\) represent joint information extracted from human and robot observations in the past $k$ timesteps.
In practice, we utilize different length k of anticipated future human joint angle at different stages, refer to 
Appendix~\ref{subsec:anticipation_explanation} for explanation.
We optimize the anticipation module parameters \(\theta\) by minimizing the prediction error:
\begin{equation}
\theta^* = \arg \min_{\theta} \sum_{\xi\in\Xi} \sum_{t=k}^{N-k} \mathcal{L}\left(\mathcal{M}_{\theta}(p_R^{t-k,t}, p_{H}^{t-k,t}), p_{H}^{t,t+k}\right),
\label{eq:model_optimization}
\end{equation}
where $\xi \in \Xi$ are buffered trajectories, and $\mathcal{L}$ is L2 Loss.

Therefore, we derive a human-aware policy for the robot,
which generates an action that more closely aligns with the user's motion tendencies based on the anticipated future joint information of the human user: 
\begin{equation}
a_{R}^{t}=\pi_R(o_R^t, \hat{p}_{H}^{t,t+k}).
\label{eq:robot_policy}
\end{equation}

In practice, our anticipation module utilizes data from the PPO buffer to predict outcomes, which, when combined with robot observations, guide the robot policy in action decisions. To maintain efficiency, the anticipation module is updated every $e_k$ epochs within the PPO policy update cycle. Importantly, the human joint information used to train the anticipation module is not shared with the PPO update algorithm, ensuring independence of information between the robot's PPO policy and the human agent.

\begin{algorithm}
\caption{Framework Pseudocode}
\begin{algorithmic}[1]
\STATE Initialize policy $\pi_R$, $\pi_H$ and estimated human utility function weight $\hat{w}_H$
\FOR{each epoch}
    \FOR{each episode}
        \STATE Collect trajectories $\{\xi_i\}$
        \STATE 
        $p_R^t$ $\gets$ $o_R^t$ and $p_H^t$ $\gets$ $o_H^t$
        \STATE Anticipation module generates predicted future joint information $\hat{p}_{H}^{t,t+k}$ with~\Cref{eq:social_motion_model}
        \STATE Policies generate actions $a_{H}$ and $a_{R}$ with~\Cref{eq:robot_policy}
        and $a_{H}^{t}=\pi_H(o_H^t)$
        
    \ENDFOR
    \STATE Calculate robot reward $r_R(\xi)$ with~\Cref{eq:robot_reward}
    and human reward $ r_H(\xi)$ with~\Cref{eq:human_reward}
    \IF{epoch mod $e_k$ = 0}
        \STATE Update Anticipation module $\mathcal{M}$ with~\Cref{eq:model_optimization}

        \STATE Update estimated human utility function weight $\hat{w}_H$ with~\Cref{eq:global_update_formula}:

    \ENDIF
    \STATE Update $\pi_R$ 
    through PPO algorithm 
    with $(\xi^D, r_R(\xi^D))$
    \STATE Update $\pi_H$ through PPO algorithm with $(\xi^D, r_H(\xi^D))$
\ENDFOR
\end{algorithmic}
\end{algorithm}

\vspace{-0.3cm}
\subsection{Utility Module}
\label{subsec:preference}

In addition to directly anticipating human behavior, it is even more crucial to recognize that human behavior is guided by an underlying utility function that reflects individual values and preferences. 
Therefore, we propose a utility module that enables the robot agent to infer an individual's utility function weight.
The utility module design is displayed in Figure \ref{fig:framework}.

Based on~\cite{palan2019learning}, the preference weight distribution can be directly obtained from expert demonstrations in the environment through a sampling process.
Acknowledging the challenge in determining utility function weights through limited demonstrations, it's noted that as the learning progresses, the policy narrows from a broad to a precise spectrum, and the consistency of demonstrated behaviors increases~\cite{kearns2002near, silver2016mastering}. 
This alignment facilitates both a reduction in the uncertainty of preference distributions and an improved ability for robots to understand human preferences as the policy converges.
Specifically, the probability of generating a specific trajectory given the utility function weights \( w \) and the current policy \( \pi \) is given by:
\begin{equation}
P(\xi^D \mid w, \mathbf{\pi}) = \prod_{i=1}^n P(\xi_i^D \mid w, \mathbf{\pi}),
\label{eq:trajectory_probability}
\end{equation}
Consequently, we can derive the global update formula, which forms the complete utility module
:
\begin{equation}
P(\hat{w}_H \mid \xi^D, t) = \exp\left(\sum_{j=1}^n \phi_{\text{pref}}(\xi_j)_{\mathbf{\pi}_t}^D\right),
\label{eq:global_update_formula_t}
\end{equation}

In reinforcement learning, the uncertainty of a policy decreases as training progresses, resulting in more stable demonstrations, which aids in estimating the utility function weights. In our approach, we aim to iteratively refine the estimated utility function weight distribution. Rather than discarding earlier estimates, we combine them linearly with the current estimates using a weighting parameter \(\gamma\). This allows us to formulate the final update rule for the utility function weight distribution as follows:
\begin{equation}
\begin{aligned}
P(\hat{w}_H \mid \xi^D) = \lim_{t \to \infty} \bigg(& \gamma \exp\left(\sum_{j=1}^n \phi_{\text{pref}}(\xi_j)_{\mathbf{\pi}_t}^D\right) \\
& + (1-\gamma) P(\hat{w}_H \mid \xi^D, t) \bigg).
\end{aligned}
\label{eq:global_update_formula}
\end{equation}

We implement utility module by initializing the utility function weights space \( \mathbf{\mathcal{X}} \) onto a unit ball, and then sample through \(N\) demonstrations $\xi^D$ under the current policy. For simplicity, we first map the high-dimensional trajectory data to a set  \( \phi_{\text{pref}}(\xi^D) \in \mathbf{\mathbb{R}}^{n*m}\) where \(n\) is number of demonstrations and \(m\) is the number of preference entries. Then, we sample the weight distribution via the Markov Chain Monte Carlo (MCMC) method~\cite{sadigh2017active}. We follow~\cite{palan2019learning} and further give a specific update formula:
\begin{equation}
f(\mathbf{\mathcal{X}}) = -\log \left( \sum e^{\mathbf{\mathcal{X}}} + (\phi_{\text{pref}}(\xi^D) \mathbf{\mathcal{X}}) \right).
\label{eq:update_formula}
\end{equation}

During training, the sampled human utility weight $\hat{w}_H$ is used to generate the estimated preference reward $\hat{r}_{\text{pref}} = \phi_{\text{pref}}(\xi; \hat{w}_H)$, which, combined with the task reward, forms the overall robot reward shaping the robot's subsequent actions:  
\begin{equation}  
r_R(\xi) = \hat{r}_\text{pref} + r_\text{task},  
\label{eq:robot_reward}  
\end{equation}  

In practice, the update frequency of this module is synchronized with the anticipation module. The process does not rely on human-provided demonstrations or queries but instead utilizes PPO buffer data, similar to the anticipation module. This enables the entire process to operate autonomously without requiring human feedback.

%% file: sections/4Experiments.tex
\section{Experiments}
\label{sec:experiment}

In this section, we evaluate our approach in collaborative human-robot interaction environments of varying robot models and assistive tasks. In particular, we seek to answer the following research questions. RQ1: What is the influence of human and robot reward misalignment? RQ2: How does our proposed method adapt to different human preferences? RQ3: How does our proposed method adapt to different assistive tasks? RQ4: How does our proposed method adapt to different robot types? RQ5: How does different merging ratio in preference weight estimation affect agent's performance?

\subsection{Setup}
\begin{figure}[t]
  \centering
  \includegraphics[width=\columnwidth]{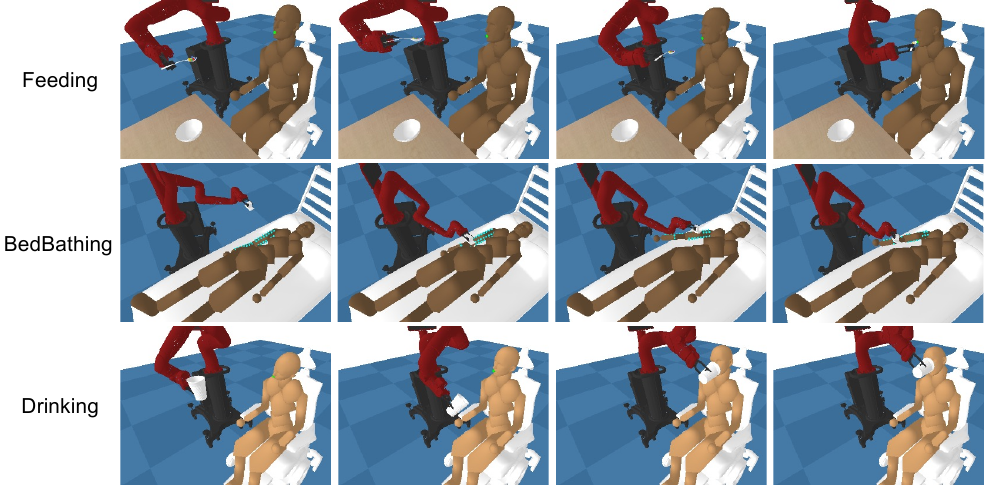}
  \caption{Successful episodes of our method for each task scenario. The key frames are arranged sequentially from left to right, following the progression of the episode.}
  \label{fig:task_visualization}
  \vspace{-5mm}
\end{figure}

\paragraph{Tasks}
We explore multiple task scenarios, each presenting unique operational challenges and distinct action patterns. In the \textit{Feeding} task, the robot is required to precisely deliver food particles to the human user's mouth using a robotic arm. The \textit{Drinking} task is similar but includes an additional complexity of arm rotation, which introduces operational difficulty. Conversely, the \textit{Bed Bathing} task differs significantly in its action pattern; it demands that the robotic arm wipe along the user's arm, maintaining constant contact and incorporating rotational movements throughout the process. 
In Figure~\ref{fig:task_visualization}, we visualize one successful episode of our method for each task. For a complete qualitative comparison with the baseline methods, please refer to the attached video.

\paragraph{Human Preference Settings}

We focus on several aspects of human preference following~\cite{erickson2020assistive}: 
\textit{Hit}, which refers to penalties incurred when particles touch points outside the target area, indicating whether the robot has delivered particles to incorrect locations or caused spillage; 
\textit{Force}, referring to the penalties for additional forces applied by the robot or tool outside the target point, representing penalties for accidental contact with non-target areas;
\textit{High Force}, which refers to penalties for excessive force applied by the robot or tool on the human body, indicative of whether the robot can adequately control the service force, typically set at 10N.
In our experiments, we evaluate our approach with simulated users, each prioritizing different preferences through varying weight combinations.

\paragraph{Metrics and Baselines} 
We compute the following metrics to comprehensively evaluate the performance of assistive policies: (1) \emph{Human Reward} \( R_H \), which includes both task success and human preference rewards per episode; (2) Decomposed human preference reward terms, capturing diffferent aspects including \emph{Hit}, \emph{Force}, and \emph{High Force} penalties per episode; and (3) Average \emph{Success Rate} across episodes.
We implement our framework on top of PPO~\cite{schulman2017proximal}. We compare it with PPO~\cite{schulman2017proximal}, TD3~\cite{fujimoto2018addressing},
and a multi-agent RL method MADDPG~\cite{lowe2017multi}. By default, we consider the scenario where human and robot rewards are misaligned. For comparison, we also experiment with a setting where the two rewards are co-optimized (\emph{co-opt.}), following~\cite{erickson2020assistive}.

\subsection{Evaluation and Comparison}

\subsubsection{Varying Human Preferences}

\begin{table}
\vspace{0.4cm}
\caption{Feeding task performance with Sawyer robot. A ``/'' denotes invalid value due to task incompletion.}
\centering
\setlength{\tabcolsep}{3pt}
\adjustbox{max width=\columnwidth}{
\footnotesize
\begin{tabular}{crrrrrr}
\toprule
\textbf{Method} & \makecell{\textbf{Human} \\ \textbf{Reward}} & \makecell{\textbf{Hit} \\ \textbf{Penalty}} & \makecell{\textbf{Force} \\ \textbf{Penalty}} & \makecell{\textbf{High} \\ \textbf{Force} \\ \textbf{Penalty}} & \makecell{\textbf{Success} \\ \textbf{Rate}} \\
\midrule
\makecell{PPO \\ (\emph{co-opt.})} & 115.992 & -0.406 & -0.756 & -2.102 & 100.0\% \\
\midrule
PPO & 109.063 & -2.668 & -1.421  & -4.645  & 98.8\%\\
TD3 & 87.069 & -2.341 & -2.149 & -12.146 & 90.0\% \\
MADDPG & -125.921 & -9.086 & -8.382 & -0.537 & 2.5\% \\
Ours & \textbf{118.656} & \textbf{-0.654} & \textbf{-0.470} & \textbf{-0.319} & \textbf{100.0\%} \\
\bottomrule
\end{tabular}
}
\label{tab:different_setting_results_averaged}
\end{table}

\begin{figure}[ht]
  \centering
  \includegraphics[width=3.1in]{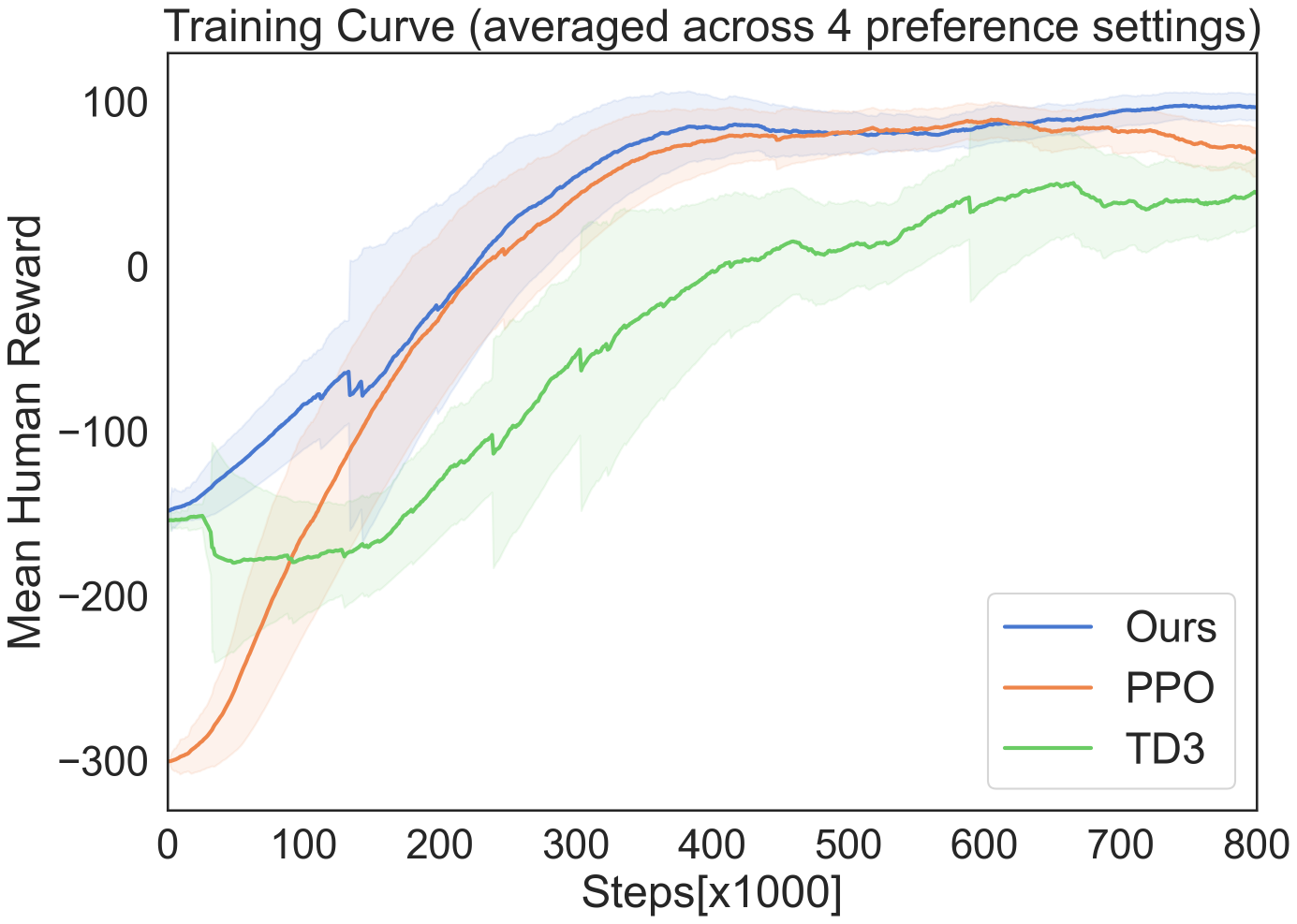}
  \caption{Training curves of baseline PPO, TD3 and our method in 4 different human preference settings (with robot Sawyer conducting the feeding task). Our method delivers generally superior performance throughout the training process.}
  \label{fig:learning_curve_4setting}
\end{figure}

In this section, we aim to address two pivotal questions: Firstly, we explore the influence of misalignment between human and robot rewards (RQ1). Secondly, we evaluate the performance of our proposed method in comparison to others among users with varying preferences (RQ2). These inquiries are fundamental to understanding the dynamics of reward systems and their impact on user satisfaction and task effectiveness in robotic applications.
We evaluate the feeding task under four different human preference settings, with results averaged across preferences shown in Table~\ref{tab:different_setting_results_averaged}. The results show that reward misalignment increases task difficulty, reducing both success rate and human reward, as seen in the first two lines.
Meanwhile, our approach steadily improves both task performance and human rewards. It not only outperforms the baseline under the same reward misalignment condition but also achieves performance comparable to or even better than the co-optimization scenario. We believe this is due to our method's ability to explicitly model the behavior of human users more effectively, enabling a better understanding of human demands.

As a reference, we present the training curves for the two primary baseline methods, PPO and TD3, alongside our own method in Figure~\ref{fig:learning_curve_4setting}.

\subsubsection{Varying Assistive Tasks}

\begin{table}[h!]
\caption{Performance on different assistive tasks. A ``/'' denotes invalid value due to task incompletion.}
\centering
\adjustbox{max width=\columnwidth}{%
\footnotesize  
\begin{tabular}{crrrrrr}
\toprule
\textbf{Method} & \makecell{\textbf{Human} \\ \textbf{Reward}} & \makecell{\textbf{Hit} \\ \textbf{Penalty}} & \makecell{\textbf{Force} \\ \textbf{Penalty}} & \makecell{\textbf{High Force} \\ \textbf{Penalty}} & \makecell{\textbf{Success} \\ \textbf{Rate}} \\
\midrule
\textbf{Feeding} \\
PPO & 99.273 & -5.500 & \textbf{-0.054} & -6.135 & 95\% \\
TD3 & 102.300 & \textbf{-2.000} & -0.324 & -6.900 & \textbf{100\%} \\
MADDPG & -213.700 & / & / & / & 0\% \\
Ours & \textbf{108.102} & \textbf{-2.000} & -0.133 & \textbf{-1.245} & \textbf{100\%} \\
\midrule
\textbf{Bed Bathing} \\
PPO & 123.585 & \textbf{0.000} & -2.886 & -2.630 & 15\% \\
TD3 & 52.372 & / & / & / & 0\% \\
MADDPG & 11.536 & / & / & / & 0\% \\
Ours & \textbf{148.558} & \textbf{0.000} & \textbf{-0.978} & \textbf{-1.204} & \textbf{60\%} \\
\midrule
\textbf{Drinking} \\
PPO & 344.189 & -53.000 & -2.359 & -0.597 & 90\% \\
TD3 & -68.404 & / & / & / & 0\% \\
MADDPG & -141.585 & / & / & / & 0\% \\
Ours & \textbf{415.390} & \textbf{-16.500} & \textbf{-0.387} & \textbf{-0.312} & \textbf{100\%} \\
\bottomrule
\end{tabular}%
}
\label{tab:different_task_results}
\end{table}

We aim to evaluate the generalization performance of our proposed method across tasks with varying operational modes. To this end, we test it on two additional complex tasks to assess whether our approach can be broadly applied to a diverse range of assistive tasks. This exploration is critical for validating the wide applicability and effectiveness of our method in various daily assistive scenarios.
The results, shown in Table~\ref{tab:different_task_results}, demonstrate strong generalization performance. For visualized comparisons, please refer to the attached video. For example, in the Bed Bathing task, TD3 fails to complete the task, and the success rate of the PPO baseline drops significantly, while our method achieves a 60\% success rate. In the Drinking task, while maintaining a 100\% task completion rate, our method reduces the Hit penalty to less than half of its original level compared to PPO.

\subsubsection{Varying Robot Types}

\begin{table}[h!]
\centering
\caption{Performance on different robots. A ``/'' denotes invalid value due to task incompletion.}
\adjustbox{max width=\columnwidth}{%
\footnotesize
\begin{tabular}{crrrrrr}
\toprule
\textbf{Method} & \makecell{\textbf{Human} \\ \textbf{Reward}} & \makecell{\textbf{Hit} \\ \textbf{Penalty}} & \makecell{\textbf{Force} \\ \textbf{Penalty}} & \makecell{\textbf{High} \\  \textbf{Force} \\  \textbf{Penalty}} & \makecell{\textbf{Success} \\ \textbf{Rate}} \\
\midrule
\textbf{Jaco} \\
PPO & 84.157 & -9.000 & -0.654 & -3.402 & 85\% \\
TD3 & 98.581 & -8.000 & -0.240 & -1.989 & \textbf{100\%} \\
MADDPG & -137.214 & / & / & / & 0\% \\
Ours & \textbf{111.074} & \textbf{-1.000} & \textbf{-0.169} & \textbf{-0.985} & \textbf{100\%} \\
\midrule
\textbf{PR2} \\
PPO & 88.850 & -2.500 & \textbf{-0.179} & -1.828 & 90\% \\
TD3 & -23.22 & -21.500 & -0.803 & -11.385 & 40\% \\
MADDPG & -97.440 & / & / & / & 0\% \\
Ours & \textbf{91.982} & \textbf{-0.000} & -0.292 & \textbf{-1.201} & \textbf{100\%} \\
\midrule
\textbf{Baxter} \\
PPO & 104.759 & -1.500 & -0.228 & -4.623 & \textbf{100\%} \\
TD3 & 100.350 & -3.500 & -0.268 & -5.602 & \textbf{100\%} \\
MADDPG & -139.963 & / & / & / & 0\% \\
Ours & \textbf{115.401} & \textbf{-0.000} & \textbf{-0.095} & \textbf{-2.074} & \textbf{100\%} \\
\midrule
\textbf{Sawyer} \\
PPO & 99.273 & -5.500 & \textbf{-0.054} & -6.135 & 95\% \\
TD3 & 102.3 & \textbf{-2.000} & -0.324 & -6.900 & \textbf{100\%} \\
MADDPG & -213.7 & / & / & -2.100 & 0\% \\
Ours & \textbf{108.102} & \textbf{-2.000} & -0.133 & \textbf{-1.245} & \textbf{100\%} \\
\bottomrule
\end{tabular}%
}
\label{tab:different_robot_results}
\end{table}

We also investigate whether the proposed method maintains its generalization ability and robustness across different robotic embodiments, a critical question for deploying to various robotic platforms in practical scenarios. Experiments are conducted using four robots—Sawyer, Baxter, Jaco, and PR2—on the feeding task, with details provided in Table~\ref{tab:different_robot_results}. 
These findings confirm the proposed method’s strong generalization ability and robustness across diverse robotic embodiments and operational scenarios.

\subsection{Ablation Studies}

\paragraph{Module Contribution}

\begin{table}[t]
\centering
\caption{Ablation results on different modules. 
}
\small
\setlength{\tabcolsep}{3pt}
\adjustbox{max width=\columnwidth}{%
\footnotesize
\begin{tabular}{crrrrr}
\toprule
\textbf{Method} & \makecell{\textbf{Human} \\ \textbf{Reward}} & \makecell{\textbf{Hit} \\  \textbf{Penalty}} & \makecell{\textbf{Force} \\ \textbf{Penalty}} & \makecell{\textbf{High} \\ \textbf{Force} \\ \textbf{Penalty}} & \makecell{\textbf{Success} \\ \textbf{Rate}} \\
\midrule
 PPO & 109.063 & -2.668 & -1.421 & -18.581 & 98.8\% \\
 \makecell{Ours \\ (\emph{w/o} Utility)} & 118.211 & -2.901 & -0.689 & -4.642 & 98.8\% \\
 Ours (full) & \textbf{118.656} & \textbf{-0.654} & \textbf{-0.470} & \textbf{-0.319} & \textbf{100.0\%} \\
\bottomrule
\end{tabular}%
}
\label{tab:dempref_ablation_results_averaged}
\end{table}

To validate the contributions of the proposed modules to task performance, we conduct a comparative analysis on the feeding task under various preference settings. As shown in Table~\ref{tab:dempref_ablation_results_averaged}, incorporating the Anticipation module alone improves overall human reward, while integrating both modules further enhances preference satisfaction and success rates, particularly in specific preference categories. This demonstrates the complementary roles and individual contributions of the two modules.

\paragraph{Distribution Merging Ratio}

\begin{table}[t]
\centering
\caption{Ablation results on merge ratio. 
}
\small
\adjustbox{max width=\columnwidth}{%
\footnotesize
\begin{tabular}{crrrrr}
\toprule
 \makecell{\textbf{Merge} \\ \textbf{Ratio}} & \makecell{\textbf{Human} \\  \textbf{Reward}} & \makecell{\textbf{Hit} \\ \textbf{Penalty}} & \makecell{\textbf{Force} \\ \textbf{Penalty}} &  \makecell{\textbf{High} \\ \textbf{Force} \\ \textbf{Penalty}} & \makecell{\textbf{Success} \\ \textbf{Rate}} \\
\midrule
0.0 & 103.668 & -2.500 & -0.198 & -2.696 & \textbf{100\%} \\
0.1 & 110.978 & -6.500 & \textbf{-0.083} & -1.427 & \textbf{100\%} \\
0.3 & 108.102 & -2.000 & -0.133 & \textbf{-1.245} & \textbf{100\%} \\
0.5 & \textbf{113.711} & \textbf{-0.500} & -0.185 & -3.608 & \textbf{100\%} \\

\bottomrule
\end{tabular}%
}
\label{tab:merge_parameters_results}
\end{table}

In Section~\ref{subsec:preference}, we describe the progressive merging process for estimating preference value combinations. To investigate the impact of different distribution merging ratios, we evaluate them in the Sawyer robot Feeding experiment. As shown in Table~\ref{tab:merge_parameters_results}, a merge ratio of 0.0 represents the complete exclusion of previously sampled distributions. Overall, the proposed merging mechanism improves human satisfaction. Overall, the proposed merging mechanism improves human satisfaction.

%% file: sections/5Conclusion.tex
\section{Conclusion}

In this work, we propose a framework that enables robots to infer human intentions and adapt to individual preferences through an anticipation module for behavior prediction and a utility module for modeling human utilities. Experiments across various robots and tasks show that the framework improves task success, efficiency, and user satisfaction.
One limitation of this work is that, due to the complexity of human subjects, experiments are primarily conducted in simulation. Future work could explore real-world deployment and incorporate explicit communication mechanisms for better human-robot collaboration.

%% file: sections/6Appendix.tex
\appendix

\label{sec:Appendix}

\subsection{Implementation Details}
We train the proposed approach on NVIDIA GeForce RTX 2080 Ti GPUs. Each experiment takes about 12 hours for Feeding task, 24 hours for bed bathing and feeding tasks.

\begin{table}[t]
\centering
\small
\caption{Preference Weights for Settings Utilized in Experiments.}
\label{tab:settings_weights}

\adjustbox{max width=\columnwidth}{%
\footnotesize
\begin{tabular}{@{}crrrrr@{}}
\toprule
\makecell{\textbf{Weight} \\ \textbf{Type}}              & \textbf{Setting 0} & \textbf{Setting 1} & \textbf{Setting 2} & \textbf{Setting 3} & \textbf{Setting 4} \\ \midrule
\makecell{\textbf{Force Non-}\\\textbf{Target Weight}}  & 0.01 & 0.01      & 0.10      & 0.001      & 0.10      \\
\makecell{\textbf{High  Forces} \\ \textbf{Weight}}       & 0.05 & 0.50       & 5.00       & 0.005       & 0.005       \\
\makecell{\textbf{Particle Hit} \\ \textbf{Weight}  }       & 1.00  & 10.0        & 1.00        & 0.10        & 10.0        \\ \bottomrule
\end{tabular}}
\vspace{-0.2cm}
\end{table}

In the experimental section~\ref{sec:experiment}, we employ four distinct preference settings~\ref{tab:settings_weights}. 
\textit{Setting 1} features High Forces Weight and Food Hit Weight, reflecting user preferences against excessive force and food spillage. \textit{Setting 2} intensifies both weights associated with the robot's force application, especially emphasizing the High Forces Weight to prevent any force on the body. \textit{Setting 3}, acting as a control group, represents users with minimal preference demands. \textit{Setting 4}, an extreme version of \textit{Setting 1}, significantly values Food Hit Weight while being less concerned with other metrics, suited for users insensitive to force but averse to food spillage.

\subsection{Anticipation Module}
\label{subsec:anticipation_explanation}

\paragraph{Dynamic Future Mechanism.}

\begin{figure}[ht]
  \centering
  \includegraphics[width=3in]{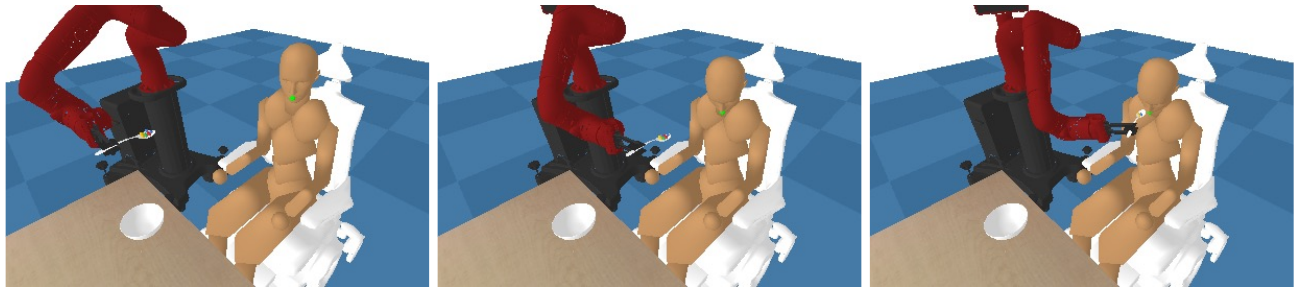}
  \caption{An example procedure of feeding task.~\textit{Left}: Early stage when Robot arm is far from human body.~\textit{Mid}: Middle stage when Robot arm is closer to human body.~\textit{Right}: Late stage when Robot arm touches human body, completing feeding. }
  \label{fig:example_feeding}
\end{figure}

In each task scenario, Assistive Gym sets the duration to 200 time steps. We demonstrate a typical task flow in the feeding task as shown in Figure~\ref{fig:example_feeding}. In the initial stage, as depicted in the left panel of Figure~\ref{fig:example_feeding}, the robot starts at a distance from the target point and needs to move the robotic arm swiftly to a position near the target. This requires anticipation of the human agent's potential movement over an extended period. Mid-task, as shown in the middle panel of Figure~\ref{fig:example_feeding}, the robot approaches the human agent while preparing to adjust its angle for delivering food, necessitating a shorter foresight than in the initial phase, yet still reliant on future predictions for accurate delivery. In the final stage, illustrated in the right panel of Figure~\ref{fig:example_feeding}, the robot's arm is close to or in contact with the target point, where long-term predictions are no longer necessary, and only immediate adjustments based on the last 1-2 time steps are required.

This example highlights that different task stages require varying lengths of future predictions. 
Accordingly, we adaptively adjust the foresight length used by the anticipation module based on the task phase, with specific adjustments as follows:
during the first 50 steps of an episode, a foresight length of 10 steps is applied; between steps 50 and 100, the foresight length is increased to 8 steps; and after 100 steps, it is further extended to 5 steps.

\subsection{Utility Module}

\paragraph{Probability Gate Design}

During our experiments, the policy learning objective prioritizes task completion first, followed by meeting human preference demands. To address this, we propose adapting the estimated preference reward for the robot agent only after achieving a certain success rate in evaluation. Specifically, we re-weight the preference estimated reward based on the latest task success rate \(\eta\).

\subsection{Training Curves}

\begin{figure}[ht]
  \centering
  \includegraphics[width=3.3in]{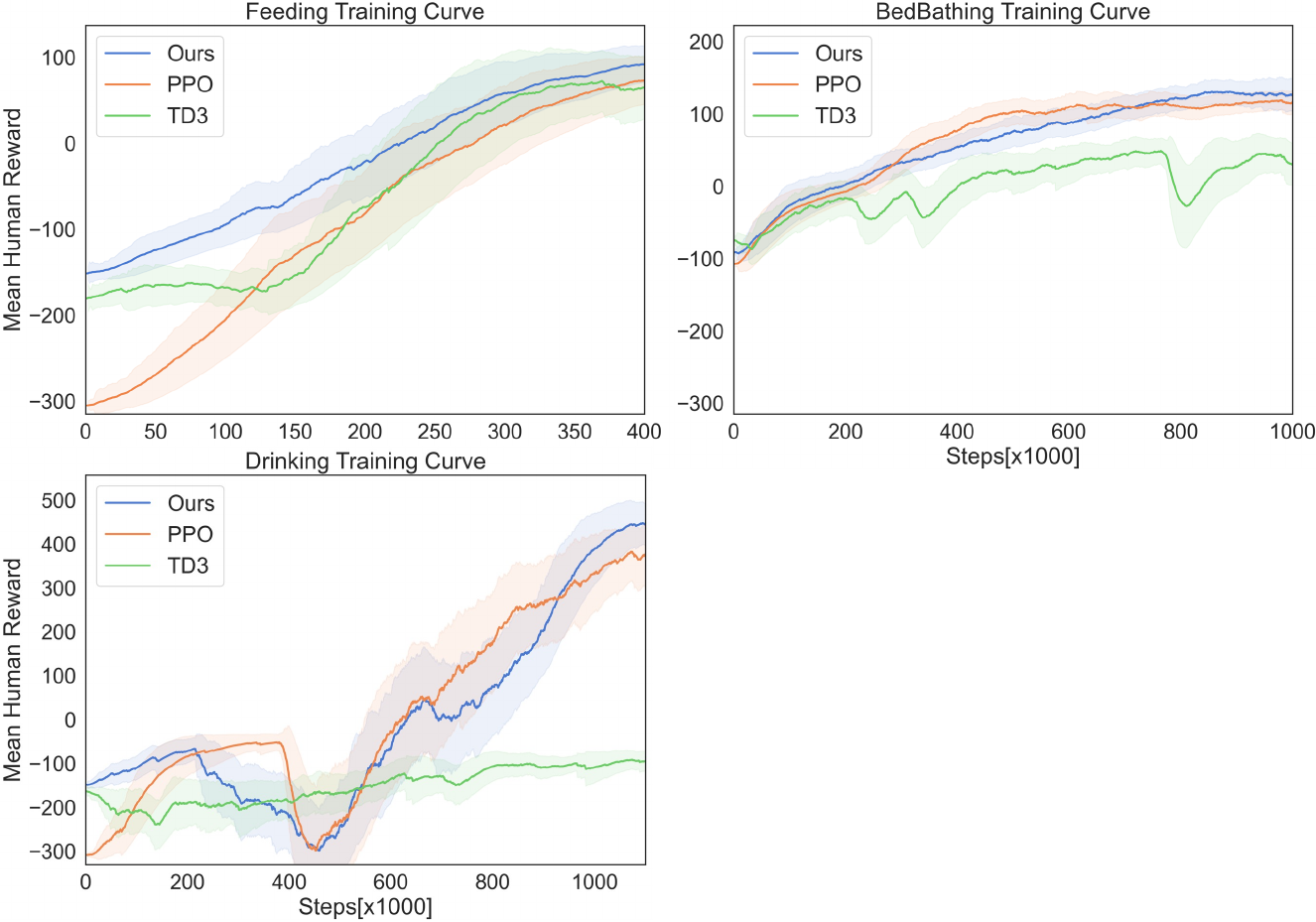}
  \caption{Training curves of baseline PPO, TD3 and our method in the feeding, bed bathing and drinking task (with robot Sawyer). Our method delivers generally superior performance throughout the training process.}
  \label{fig:learning_curve_3task}
\end{figure}

\begin{figure}[ht]
  \centering
  \includegraphics[width=3.3in]{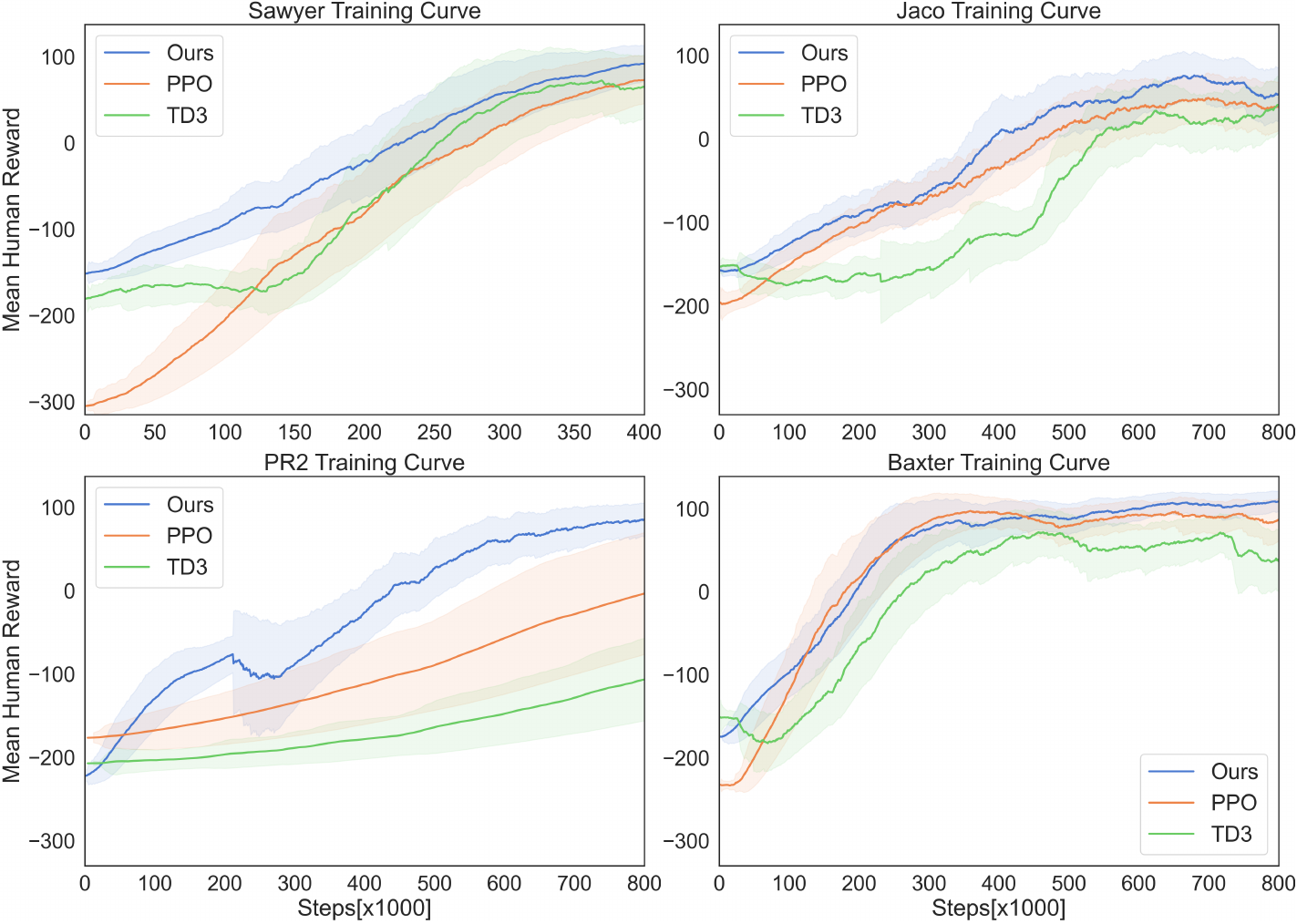}
  \caption{Training curves of baseline PPO, TD3 and our method in 4 different robot scenarios in the feeding task. Our method consistently 
 delivers superior performance throughout the entire training process.}
  \label{fig:learning_curve_4robot}
\end{figure}

We provide additional training curves to compare the performance of different methods across various scenarios.
Figure~\ref{fig:learning_curve_3task} shows the training curves for different assistive tasks, where our method achieves superior performance throughout training. Figure~\ref{fig:learning_curve_4robot} presents the training curves for different robots performing the feeding task, demonstrating the stability and consistent superiority of our method across the training process.

